\documentclass{article}
\usepackage{ijcai23}

\usepackage{times}
\usepackage{soul}

\usepackage{amsmath,amsfonts,bm}

\def\eqref#1{equation~\ref{#1}}

\def\1{\bm{1}}

\def\vh{{\bm{h}}}

\def\vk{{\bm{k}}}

\def\vq{{\bm{q}}}

\def\vv{{\bm{v}}}

\def\vx{{\bm{x}}}

\def\mA{{\bm{A}}}

\def\mF{{\bm{F}}}

\def\mH{{\bm{H}}}

\def\mK{{\bm{K}}}

\def\mP{{\bm{P}}}
\def\mQ{{\bm{Q}}}

\def\mV{{\bm{V}}}

\DeclareMathAlphabet{\mathsfit}{\encodingdefault}{\sfdefault}{m}{sl}
\SetMathAlphabet{\mathsfit}{bold}{\encodingdefault}{\sfdefault}{bx}{n}

\newcommand{\R}{\mathbb{R}}

\newcommand{\softmax}{\mathrm{softmax}}

\DeclareMathOperator*{\argmax}{arg\,max}

\usepackage{url}
\usepackage[hidelinks]{hyperref}
\usepackage[utf8]{inputenc}
\usepackage[small]{caption}
\usepackage{xcolor}
\usepackage{graphicx}
\usepackage{amsmath}
\usepackage{float}
\usepackage{amsthm}
\usepackage{booktabs}
\usepackage{algorithm}
\usepackage{tikz}
\usepackage{algorithmic}
\usepackage{pgfplots}
\usepackage{pgfplotstable}
\usepackage{adjustbox}
\usepackage{multirow}
\usepackage{multicol}
\usepackage{caption}
\usepackage{subcaption}
\usepackage[switch]{lineno}

\author{
Wenhao Zhu$^1$
\and
Guojie Song\footnote{Corresponding Author}$^2$\and
Liang Wang$^3$\And
Shaoguo Liu$^4$\\
\affiliations
$^{1,2}$National Key Laboratory of General Artificial Intelligence, School of Intelligence Science and Technology, Peking University\\
$^{3,4}$Alibaba Group\\
\emails
\{wenhaozhu, gjsong\}@pku.edu.cn,
\{liangbo.wl, shaoguo.lsg\}@alibaba-inc.com,
}

\newtheorem{definition}{Definition}[section]
\newtheorem{fact}{Fact}
\newcommand*{\ldblbrace}{\{\mskip-5mu\{}
\newcommand*{\rdblbrace}{\}\mskip-5mu\}}

\begin{document}

\title{AnchorGT: Efficient and Flexible Attention Architecture for Scalable Graph Transformers}




\maketitle

\begin{abstract}
  Graph Transformers (GTs) have significantly advanced the field of graph representation learning by overcoming the limitations of message-passing graph neural networks (GNNs) and demonstrating promising performance and expressive power. However, the quadratic complexity of self-attention mechanism in GTs has limited their scalability, and previous approaches to address this issue often suffer from expressiveness degradation or lack of versatility. To address this issue, we propose AnchorGT, a novel attention architecture for GTs with global receptive field and almost linear complexity, which serves as a flexible building block to improve the scalability of a wide range of GT models. Inspired by anchor-based GNNs, we employ structurally important $k$-dominating node set as anchors and design an attention mechanism that focuses on the relationship between individual nodes and anchors, while retaining the global receptive field for all nodes. With its intuitive design, AnchorGT can easily replace the attention module in various GT models with different network architectures and structural encodings, resulting in reduced computational overhead without sacrificing performance. In addition, we theoretically prove that AnchorGT attention can be strictly more expressive than Weisfeiler-Lehman test, showing its superiority in representing graph structures. Our experiments on three state-of-the-art GT models demonstrate that their AnchorGT variants can achieve better results while being faster and significantly more memory efficient.
\end{abstract}

\section{Introduction}
\label{sec1}
Transformer \cite{vaswani2017attention} has become the dominant universal neural architecture for natural language processing and computer vision due to its powerful self-attention mechanism. Its success has sparked interest in adapting Transformer for use in graph machine learning \cite{ying2021transformers,rampavsek2022recipe}. While Graph Neural Networks (GNNs) have limitations such as over-smoothing and neighbor explosion in the message-passing paradigm, the promising performance of Transformer-based approaches in graph machine learning has led researchers to investigate their potential use in a wider range of scenarios.

However, the application of Transformer in large-scale graph machine learning is often limited by the $O(N^2)$ complexity of the self-attention mechanism, where $N$ is the number of nodes in the input graph. To address this computational bottleneck, previous approaches have typically employed two strategies: restricting the receptive field of nodes through techniques such as sampling \cite{zhang2022hierarchical}, or applying linear attention methods directly to graph Transformer \cite{rampavsek2022recipe}. However, both of these approaches have inherent flaws. The first strategy sacrifices the key element of self-attention - global receptive field - which reduces the model's ability to capture global graph structure and may ignore graph structural information with simple sampling methods. The second approach, while maintaining the global attention mechanism, is incompatible with the common relative structural encoding in graph Transformer (as the approximate step bypasses computation of the full attention matrix, will be discussed later), significantly reducing the model's ability to learn graph structure. As a result, current approaches to improving the scalability of graph Transformer sacrifice global receptive field or structural expressivity to some extent and still require improvement.

To address these issues, we draw inspiration from anchor-based graph neural networks \cite{you2019position} and propose AnchorGT, a novel attention architecture for graph Transformers with almost linear complexity, high flexibility, and strong expressive power. Previous anchor-based graph neural networks often use a randomly sampled set of anchor nodes as a tool for GNNs to capture the global relative positions of nodes in a graph. In our approach, we select a set of $k$-dominating set nodes, which are nodes that can be quickly computed and are structurally important, as the anchor set to support the learning and propagation of broad-area information in the graph structure. Based on these anchors, our redesigned attention mechanism allows each node to attend to both its neighbors and the anchor nodes in the global context, effectively reducing computational complexity while retaining the fundamental characteristics of the attention mechanism. With the $k$-dominating set anchors and the new attention mechanism, AnchorGT layer achieves almost linear complexity, has global receptive field for each node, and is compatible with many structural encodings and graph Transformer methods. Moreover, we theoretically prove that the AnchorGT layer with structural encoding that satisfies certain conditions is strictly more expressive than graph neural network based on Weisfeiler-Lehman test, further demonstrating the superiority of our method.

In experiments, we replaced the attention mechanism module of three state-of-the-art graph transformer methods with anchor-based attention and tested their performance and scalability on both graph-level and node-level datasets. The results show that the AnchorGT variants of these methods have significantly reduced memory consumption and training time while maintaining excellent performance, proving that AnchorGT can effectively improve the scalability of graph Transformers without sacrificing performance. We summarize the contribution of this paper as follows:

\begin{itemize}
    \item We propose AnchorGT, a novel attention architecture for graph Transformers as a flexible building block to improve the scalability of a wide range of graph Transformer models.
    \item In AnchorGT, we propose a novel approach of using $k$-dominating set anchors to efficiently propagate global graph information, and design a new attention mechanism that combines local and global information based on the anchor method. AnchorGT has almost linear complexity and global receptive field, and can be integrated with a wide range of existing graph Transformer methods.
    \item Theoretically, we prove that AnchorGT layer can be strictly more expressive than WL-GNNs with certain structural encoding.
    \item In experiments, the AnchorGT variants of three state-of-the-art graph Transformer models achieved competitive performance on both graph-level and node-level tasks, with significantly improved scalability.
\end{itemize}

\begin{figure*}[t]
\centering
\includegraphics[width=0.85\linewidth]{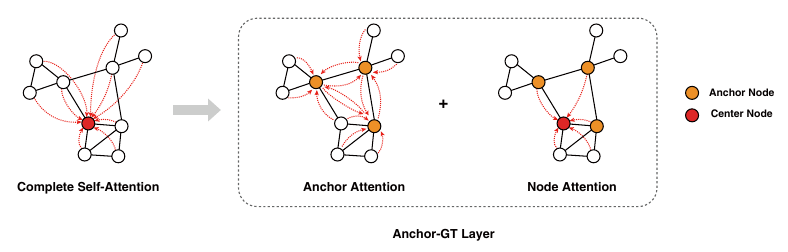}
\caption{An illustration of the proposed AnchorGT.}
\label{fig1}
\end{figure*}

\section{Related Work}

\subsection{Graph Transformer}
Along with the recent surge of Transformer, many prior works have attempted to bring Transformer architecture to the graph domain, including GT \cite{dwivedi2020generalization}, GROVER \cite{rong2020self}, Graphormer \cite{ying2021transformers}, SAN \cite{kreuzer2021rethinking}, SAT \cite{chen2022structure}, ANS-GT \cite{zhang2022hierarchical}, GraphGPS \cite{rampavsek2022recipe}, GRPE \cite{park2022grpe}, EGT \cite{hussain2022global}, NodeFormer \cite{wunodeformer}, GOAT \cite{pmlr-v202-kong23a}. \cite{muller2023attending} is an complete survery of graph Transformers. GT \cite{dwivedi2020generalization} provides a generalization of Transformer architecture for graphs with modifications like using Laplacian eigenvectors as positional encodings and adding edge feature representation to the model. GROVER \cite{rong2020self} is a molecular large-scale pretrain model that applies Transformer to node embeddings calculated by GNN layers. Graphormer \cite{ying2021transformers} proposes an enhanced Transformer with centrality, spatial and edge encodings, and achieves state-of-the-art performance on many molecular graph representation learning benchmarks. SAN \cite{kreuzer2021rethinking} presents a learned positional encoding that cooperates with full Laplacian spectrum to learn the position of each node in the graph. Gophormer \cite{zhao2021gophormer} applies structural-enhanced Transformer to sampled ego-graphs to improve node classification performance and scalability. SAT \cite{chen2022structure} studies the question of how to encode structural information to Transformers and proposes the Structure-Aware-Transformer to generate position-aware information for graph data. ANS-GT \cite{zhang2022hierarchical} proposes an adaptive sampling strategy to effectively scale up graph Transformer to large graphs. GraphGPS \cite{rampavsek2022recipe} proposes a general, powerful and scalable architecture for building graph Transformers and use the fast attention method to improve scalability. GOAT \cite{pmlr-v202-kong23a} proposes a global graph Transformer architecture with excellent performance and efficiency. NodeFormer \cite{wunodeformer} introduces a novel all-pair message passing
scheme for efficiently propagating node signals between arbitrary nodes based on the Gumbel-Softmax operator.

\subsection{Anchor-based Graph Neural Network}
Anchor-based Graph Neural Network is first proposed by P-GNN \cite{you2019position}. To generate position-aware node embeddings that captures the positional information of nodes within the global context, P-GNN proposes to incorporate distance information from the target node to a set of randomly sampled anchor nodes in the whole graph during the representation generation process. Following P-GNN, AS-GNN \cite{dong2021improving} proposes the MVC algorithm, utilizing complex network theory, to select structurally important nodes based on subgraph partitioning as anchor nodes to enhance the performance of GNN representations of graph structure, instead of random selection. IDGL \cite{chen2020iterative} proposes an graph learning framework for jointly and iteratively learning graph structure and graph embedding with anchor-based approximation technique to reduce computational complexity. A-GNN \cite{liu2020gnn} selects anchors through greedy algorithm with the minimum point cover algorithm, and learns a non-linear aggregation scheme to incorporate the anchor-based structural information into message-passing scheme. Previous anchor-based graph neural network methods have explored various ways of incorporating anchor information to improve graph neural networks, and our work is the first to apply anchor-based methods to graph Transformer, resulting in a efficient and flexible new graph Transformer framework.

\section{Proposed Approach}

In this section, we will describe the overall architecture of the model, including definition of anchor sets and the anchor-based attention module.

\paragraph{Notations}
Let $G=(V,E)$ denote a graph, where $V=\{v_1,v_2,\ldots,v_n\}$ is the node set that consists of $n$ vertices and $  E\subset   V\times   V$ is the edge set. For $v,u\in V$, let $\text{SPD}(v,u)$ be the shortest path distance between $v$ and $u$ in $G$. For node $v\in  V$, let $  N(v)=\{v':v'\in  V, (v,v')\in  E\}$ denote the set of its neighbors, and let $N_k(v)=\{u:u\in V,\text{SPD}(v,u)\leq k\}$ be the $k-$hop neighborhood of $v$. Let each node $v_i$ be associated with a feature vector $\vx_i\in \mathbb{R}^F$ where $F$ is the hidden dimension, and let $\mathbf{X}=[\vx_1,\vx_2,\ldots,\vx_n]^\top\in\mathbb{R}^{n\times F}$ denote the feature matrix.

\subsection{$k-$Dominating Set Anchor}

In graphs, nodes and edges often possess unique structural positions, where some nodes occupy central positions in the overall structure (such as hub nodes in social network graphs \cite{wasserman1994social} and key carbon atoms in molecular graphs), while others are situated in peripheral locations. Studies on complex network dynamics \cite{chen2012identifying} have also demonstrated that certain key nodes and edges in real-world graphs often possess higher information propagation probabilities than other nodes, and these nodes generally contain more significant global structural information. Therefore, we can select a subset of nodes that possess key structural information as anchors, making them an efficient medium for conveying global information in the graph.

In the proposed AnchorGT, we choose the $k-$dominating set ($k-$DS) as the anchor set, as this type of anchor set can be calculated with approximate linear complexity and has structural significance. For a $k-$DS anchor set $S$, any node $v\in V$ is an anchor node or there exists an anchor $u\in S$ such that $\text{SPD}(v,s)\leq k$. Formally, the $k-$dominating set is a classical concept in graph theory, defined as follows:

\begin{definition}[$k-$dominating set]
For graph $G=(V,E)$, its $k-$dominating set $S$ is a subset of $V$ such that for any $v\in V$, there exists $u\in S$ satisfying $\text{SPD}(v,u)\leq k$.
\end{definition}

In order to compute the $k-$dominating set, we follow previous studies \cite{nguyen2020solving} and adopt an approximate linear complexity greedy algorithm. The specific steps of the algorithm are described in Algorithm 1 in appendix. This algorithm first assigns labels to all nodes in the graph, and repeatedly selects the labeled node with the highest degree from the graph and adds it to the anchor set, while removing labels from its $k-$hop neighborhood, until all nodes are unlabeled. The complexity of degree sorting is $O(N\log N)$. The algorithm performs at most $N$ iterations with at most $n_k$ queries of the adjacency matrix in each iteration, where $n_k$ is the maximum number of $k-$hop neighborhood nodes. Therefore, the overall complexity of the algorithm is $O(N(\log N+n_k))$, which is an approximate linear complexity with respect to $N$ since $\log N\ll N$ and $n_k\ll N$. In the experiment, the time required to compute the 2-DS of the ogbn-products dataset with over 3 million nodes is less than 20 minutes, which is negligible compared to the training time of the network and demonstrates that the computational complexity of this part is completely acceptable. The $k-$dominating set $S$ possesses a key property - the union of the $k-$hop neighborhoods of all nodes in the set results in the complete node set $V$, which will be crucial for our model to achieve global perception.

\subsection{Anchor-based Attention}
In the following, we will describe the proposed anchor-based attention mechanism. In this method, each node will perform attention calculation on both its $k-$hop neighbors and all anchor nodes, to achieve the goal of reducing complexity and maintaining global receptive field. Specifically, for every $v\in V$ with representation $\vh_v$, its AnchorGT self-attention computation follows the following scheme:

\begin{align}
    &\vq_v=\mP_Q(\vh_v), \\
    &\text{for }u\in R(v):\\
    &\quad\vk_u=\mP_K(\vh_u),\vv_u=\mP_V(\vh_u),\\
    &\quad\alpha_{v,u}=\softmax_{u\in R(v)}(\frac{\vq_v^\top\vk_u}{\sqrt d}+\mF(\text{SE}(v,u))),
\end{align}
where $\mP_Q,\mP_K,\mP_V$ are learnable linear projection functions for query, key, value vectors, $\mF$ is the transform function on structural encodings (e.g. table embedding or linear transform), and $R(v)$ is the \textit{receptive field} of $v$ which we will use to control the scope of attention mechanism. $\text{SE}(v,u)$ is the \textit{structural encoding} of node pair $(v,u)$ to inject structural information into self-attention. The output representation is 
\begin{align}
\vh_v=\sum_{u\in R(v)}\alpha_{v,u}\vv_u.
\end{align}
Then in anchor-based attention, with $k-$DS anchor set $S$, the receptive field of nodes is defined as
\begin{align}
R(v)=N_k(v)\cup S.
\end{align}

The above definition constitutes our proposed Anchor-based attention mechanism in the AnchorGT framework, which is independent of specific structural encoding and can easily be extended to multi-head attention scenarios. It can be seamlessly incorporated into various graph Transformer networks to replace the original attention mechanism. As all the $k-$hop neighbors of the anchor points cover the entire node set, we can conclude that after more than one rounds of attention computation, every node has a global receptive field. Moreover, assuming $A$ is the size of the anchor set $S$, we can easily infer that the computational complexity of the model is $O(N(n_k+A))$. Since $n_k\ll N$ and $A\ll N$ (which we will prove by numbers in Section \ref{sec64}), the complexity of Anchor-based attention is also nearly linear. As a result, our proposed model is theoretically more scalable than previous methods with $O(N^2)$ attention mechanism, and it has been shown by experiments to result in significantly lower GPU memory usage and less training time. We primarily utilized the \texttt{torch.sparse.sampled\_addmm} function from PyTorch package to implement the proposed anchor-based attention module.

\paragraph{Structural Encoding} The vanilla self-attention mechanism is ignorant of the graph structure, so it is necessary to incorporate structural encoding to learn graph structural information. Common methods include absolute structural encoding (including node degree \cite{ying2021transformers}, Laplacian eigenvectors \cite{kreuzer2021rethinking}) and relative structural encoding (including shortest path length and transition probability \cite{rampavsek2022recipe}). Our model is not dependent on a specific structural encoding definition. In experiments, we adopt the shortest path structural (SPD) encoding \cite{ying2021transformers}, which is simple and effective, and can be calculated in linear time. In subsequent theoretical analysis, we will also prove that this encoding makes the model have stronger expressive power than graph neural networks.

\paragraph{Anchor-based Attention with Sampling-based Graph Transformers}
In the previous paragraphs, we describe the anchor-based attention mechanism in graph Transformer models based on the entire graph. This type of model typically processes multiple small-scale graphs (such as molecular graphs) in batches and predicts node or graph labels, like Graphormer \cite{ying2021transformers} and GraphGPS \cite{rampavsek2022recipe} described in the original papers. However, large-scale graph data exists in the real world, which generally involves node-level or edge-level tasks and is mostly trained using sampling-based training methods, like ANS-GT \cite{zhang2022hierarchical}. In sampling-based training, each training batch samples smaller sub-graph that can be accommodated by the GPU memory from the global large graph, and then the model calculates the node or edge representations on the sub-graph. We can see that our method is also applicable to this type of graph Transformer. Our model only needs to pre-calculate the anchor set of the entire graph, then add the subset of the graph anchor set in the sampled sub-graph to each batch during training.

\section{Expressiveness of AnchorGT}
\label{sec5}

Previous results have demonstrated that graph Transformers may possess stronger expressive power than GNNs, and our model can be considered as a simplified one with randomness. Therefore, a natural question arises: can we provide similar theoretical guarantees for the proposed AnchorGT model? In the following section, we will give a affirmative answer to this question - as long as structural encodings that satisfy certain conditions (such as SPD) are utilized, our model can have strictly stronger expressive power than GNNs. In this section, we consider a simple AnchorGT model built by stacked Transformer layers with AnchorGT attention and a global readout function (like mean in GNNs).

Considering that the graph structure information in graph Transformer models is entirely provided by the structural encoding function, we must first impose certain requirements on these structural encodings to ensure a lower bound on the model's expressive power. We first define the \textbf{neighbor-distinguishable} structural encoding:
\begin{definition}[Neighbor-Distinguishable Structural Encoding]
In $G=(V,E)$, a relative structural encoding function $\text{SE}(\cdot,\cdot)\in V\times V\mapsto\R^d$ is neighbor-distinguishable if a mapping $f:\R^d\mapsto\{0,1\}$ exists such that for $v\in V$ and $u\in R(v)$, $f(\text{SE}(v,u))=1$ if $(v,u)\in E$, and $f(\text{SE}(v,u))=0$ if $(v,u)\notin E$. 
\end{definition}

For example, the shortest-path-distance structural encoding is neighbor-disguishable since two nodes are adjacent if their SPD is 1. Using the similar method previously outlined in the literature \cite{ying2021transformers} to simplify graph Transformer models into GNN models, we can easily prove the following conclusion:

\begin{fact}
For any two graphs $G_1$ and $G_2$ that can be distinguished by a message-passing GNN model, there exists a set of model parameters such that the AnchorGT model with neighbor-distinguishable encoding can distinguish them.
\end{fact}

Noting that the aggregation function and global readout function in GNNs can both be expressed by graph Transformer, the proof of the above conclusion is intuitive. We provide a detailed proof in the appendix. The above conclusion demonstrates that as long as there is a neighbor-distinguishable encoding, the AnchorGT model's expressive power is not weaker than GNNs. However, what we are most concerned with is: \textit{Does AnchorGT allow the graph Transformer to surpass the expressiveness of GNNs?} To answer this question, we first provide the definition of \textit{anchor-distinguishable} structural encoding:

\begin{definition}[Anchor-Distinguishable Structural Encoding]
In $G=(V,E)$ with $k-$DS anchor $S\subset V$, a relative structural encoding function $\text{SE}(\cdot,\cdot)\in V\times V\mapsto\R^d$ is anchor-distinguishable if a mapping $f:\R^d\mapsto\{0,1\}$ exists such that for $v\in V$ and $u\in R(v)$, $f(\text{SE}(v,u))=1$ if $u\in S$ and $u\notin N_k(v)$, and $f(\text{SE}(v,u))=0$ otherwise. 
\end{definition}

Anchor-distinguishable structural encodings allow the model to access structural information from anchors outside the neighborhood, which is crucial for expressive power. \textit{The SPD encoding is anchor-distinguishable} for any $k$, because the SPD encoding only produces results $\geq k$ for anchor nodes outside of $k$-neighbors. We prove the following result (detailed proof in appendix):

\begin{fact}
\label{fact2}
There exist graphs $G_1$ and $G_2$ that can not be distinguished by message-passing GNNs, but can be distinguished by AnchorGT model with $1-$DS anchors and structural encoding that is both neighbor-distinguishable and anchor-distinguishable.
\end{fact}

The above conclusion theoretically proves that as long as the structural encoding is neighbor-and-anchor-distinguishable (like SPD we use), our AnchorGT model can have strictly stronger expressive power than GNNs; this expressive power comes from the combination of the anchor method and strong structural encoding. Interestingly, under the SPD structural encoding, AnchorGT model can be seen as a simplified version of Graphormer, but can still distinguish $G_1$ and $G_2$ (Figure 4 in Appendix) which serve as an example to prove that Graphormer has an expressive ability superior to GNNs \cite{ying2021transformers}. Therefore, the AnchorGT method allows attention mechanism to maintain low complexity while maintaining the expressive power advantage of Graph Transformer compared to GNNs.

\section{Experiments}
In experiments, we test the proposed AnchorGT method on three graph Transformer models and both graph-level and node-level benchmarks. We show that the AnchorGT variants of graph Transformer models can achieve the same performance level as the original models while being significantly more efficient. 

\begin{table*}[h]
    \begin{adjustbox}{width=1.90\columnwidth,center}
    \begin{tabular}{l|ccccc}
        \toprule
        Dataset & ZINC & QM9 & QM8 & LRGB-Peptides-struct & LRGB-Peptides-func \\
        \midrule
        Metric & MAE$\downarrow$ & Multi-MAE$\downarrow$ & Multi-MAE$\downarrow$ & MAE$\downarrow$ & AP$\uparrow$ \\
        \midrule
        GCN       & 0.1379  & 1.006\small$\pm$0.020 & 0.0279\small$\pm$0.0001 & 0.3496\small$\pm$0.0013 & 0.5930\small$\pm$0.0023 \\
        GCNII     & -       & 0.945\small$\pm$0.462 & 0.0256\small$\pm$0.0002 & 0.3471\small$\pm$0.0010 & 0.5543\small$\pm$0.0078 \\
        GAT       & -       & 1.112\small$\pm$0.018 & 0.0317\small$\pm$0.0001 & -                       & -                       \\
        GIN (GINE) & 0.1218  & 1.225\small$\pm$0.055 & 0.0276\small$\pm$0.0001& 0.3547\small$\pm$0.0045 & 0.5498\small$\pm$0.0079 \\
        GatedGCN  & -       & -                     & & 0.3420\small$\pm$0.0013 & 0.5864\small$\pm$0.0077 \\
        \midrule
        GT        & 0.226\small$\pm$0.01             & - & - & - & - \\
        SAT       & -   & 1.122\small$\pm$0.135 & 0.0281\small$\pm$0.0025 & 0.6473\small$\pm$0.0849 & 0.2601\small$\pm$0.0248 \\
        SAN (LapPE)& 0.139\small$\pm$0.006            & - & - & 0.2683\small$\pm$0.0043 & 0.6384\small$\pm$0.0121 \\
        \midrule
        Graphormer          & \textbf{0.120}\small$\pm$0.008 & 0.720\small$\pm$0.035          & 0.0079\small$\pm$0.0001          & 0.2705\small$\pm$0.0045 & 0.6453\small$\pm$0.0121 \\
        Graphormer-AnchorGT-SPD & 0.122\small$\pm$0.006 & \textbf{0.702}\small$\pm$0.051 & \textbf{0.0077}\small$\pm$0.0001 & \textbf{0.2519}\small$\pm$0.0033          & \textbf{0.6520}\small$\pm$0.0264 \\
        \midrule
        $\Delta$ & $\textcolor{blue}{-1.66\%}$ & $\textcolor{red}{+2.56\%}$ & $\textcolor{red}{+2.53\%}$ & $\textcolor{red}{+7.38\%}$ & $\textcolor{red}{+1.04\%}$ \\
        \bottomrule
    \end{tabular}
    \end{adjustbox}
    \caption{Results of Graphormer and its variant Graphormer-AnchorGT on graph representation learning benchmarks.}
    \label{tbl1}
    
\end{table*}

\begin{table*}[h]
    \begin{adjustbox}{width=1.90\columnwidth,center}
    \begin{tabular}{l|ccccc}
        \toprule
        Dataset & ogb-PCQM4Mv2 & QM9 & LRGB-Peptides-struct & LRGB-Peptides-func & CLUSTER \\
        \midrule
        Metric & Validation MAE$\downarrow$ & MAE$\downarrow$ & MAE$\downarrow$ & AP$\uparrow$  & Accuracy$\uparrow$ \\
        \midrule
        GCN       & 0.1379  & 1.006\small$\pm$0.020 & 0.3496\small$\pm$0.0013 & 0.5930\small$\pm$0.0023 & 68.498\small$\pm$0.976 \\
        GCNII     & -       & 0.945\small$\pm$0.462 & 0.3471\small$\pm$0.0010 & 0.5543\small$\pm$0.0078 & - \\
        GAT       & -       & 1.112\small$\pm$0.018 & -                       & -                       & 70.587\small$\pm$0.447 \\
        GIN (GINE) & 0.1218  & 1.225\small$\pm$0.055 & 0.3547\small$\pm$0.0045 & 0.5498\small$\pm$0.0079 & 64.716\small$\pm$1.553 \\
        GatedGCN  & -       & -                     & 0.3420\small$\pm$0.0013 & 0.5864\small$\pm$0.0077 & 73.840\small$\pm$0.326 \\
        \midrule
        SAT            & -      & 1.122\small$\pm$0.135 &  0.6473\small$\pm$0.0849 & 0.2601\small$\pm$0.0248 & 77.856\small$\pm$0.104 \\
        SAN (LapPE)     & -      & -                     & 0.2683\small$\pm$0.0043 & 0.6384\small$\pm$0.0121 & 76.691\small$\pm$0.65 \\
        Graphormer     & 0.0864 & 0.729\small$\pm$0.044 & 0.2705\small$\pm$0.0045 & 0.6453\small$\pm$0.0121  & 77.552\small$\pm$0.546 \\
        \midrule
        GraphGPS            & 0.0878 & \textit{0.705}\small$\pm$0.081 & 0.2500\small$\pm$0.0005 & \textbf{0.6535}\small$\pm$0.0041 & 78.016\small$\pm$0.180 \\
        GraphGPS-AnchorGT     & 0.0897          & 0.712\small$\pm$0.074          & 0.2674\small$\pm$0.0011          &       0.6214\small$\pm$0.0056          & \textit{78.123}\small$\pm$0.365 \\
        GraphGPS-AnchorGT-SPD & \textbf{0.0862} & \textbf{0.665}\small$\pm$0.086 & \textbf{0.2437}\small$\pm$0.0006 & \textit{0.6460}\small$\pm$0.0031 & \textbf{78.802}\small$\pm$0.252\\
        \midrule
        $\Delta$ & $\textcolor{red}{1.86\%}$ & $\textcolor{red}{+5.6\%}$ & $\textcolor{red}{+2.58\%}$ & $\textcolor{blue}{-1.15\%}$ & $\textcolor{red}{+1.01\%}$\\
        \bottomrule
    \end{tabular}
    \end{adjustbox}
    \caption{Results of GraphGPS and its AnchorGT variants on graph representation learning benchmarks.}
    \label{tbl2}
    
\end{table*}

\begin{table*}[t]
    \centering
    \begin{adjustbox}{width=1.65\columnwidth,center}  
    \begin{tabular}{c|ccccccc}
    \toprule
    Dataset       & $N$ & $E$ & avg. degree & $|A| (k=1)$ & $|A| (k=2)$ & $|A| (k=3)$ & $|A| (k=4)$\\
    \midrule
    QM9           & 2449029 & 121318152 & 25.2  & OOM   & 80970 & 60079 & 50436 \\
    ogbn-products & 18.03   & 18.66     & 1.03  & 6.8 & 3.5   & 2.25  & 1.55  \\
    \bottomrule
    \end{tabular}
    \end{adjustbox}
    \vspace{4mm}
    \caption{Statistics and number of anchors under different $k$ values of qm9 and ogbn-products dataset.}
    \label{tbl4}
\end{table*}

\begin{table}[h]  
\centering  
    \begin{adjustbox}{width=0.95\columnwidth,center}  
    \begin{tabular}{l|ccc}  
        \toprule  
        Dataset & Citeseer & ogbn-arxiv & ogbn-products \\  
        \midrule  
        Metric & \multicolumn{3}{c}{Accuracy$\uparrow$}  \\  
        \midrule  
        GCN             & 79.43\small$\pm$0.26 & 71.24\small$\pm$0.29 & 75.64\small$\pm$0.21 \\  
        GAT             & 80.13\small$\pm$0.62 & 57.88\small$\pm$0.18 & 79.45\small$\pm$0.59 \\  
        GraphSAGE       & 79.23\small$\pm$0.53 & 71.49\small$\pm$0.27 & 78.50\small$\pm$0.14 \\  
        APPNP           & 79.33\small$\pm$0.35 & - & - \\  
        \midrule  
        SAN             & 70.64\small$\pm$0.97 & - & - \\  
        Gophormer       & 76.43\small$\pm$0.78 & - & - \\  
        \midrule  
        ANS-GT          & \textbf{80.25}\small$\pm$0.39 & 71.86\small$\pm$0.20 & \textbf{80.50}\small$\pm$0.78 \\  
        ANS-GT-AnchorGT & 82.01\small$\pm$0.21 & \textbf{72.34}\small$\pm$0.45 & 81.04\small$\pm$0.52 \\  
        \midrule  
         $\Delta$ & $\textcolor{red}{+2.19\%}$ & $\textcolor{red}{+0.67\%}$ & $\textcolor{red}{+0.67\%}$\\  
        \bottomrule  
    \end{tabular}  
    \end{adjustbox}  
    \caption{Results of node representation learning benchmarks.}  
    \label{tbl3}  
\end{table}

\subsection{Datasets and Experimental Settings}

In graph representation learning experiments, we select 7 datasets: ogb-PCQM4Mv2 from ogb-lsc \cite{hu2020open,hu2021ogb}, QM9 and QM8 from MoleculeNet \cite{wu2018moleculenet}, LRGB-Peptides-struct and LRGB-Peptides-func from LRGB datasets \cite{dwivedi2022long}, and CLUSTER \cite{dwivedi2020benchmarking}. In node representation learning experiments, we select 4 datasets: Citeseer, Pubmed \cite{kipf2016semi}, ogbn-arxiv and ogbn-products \cite{hu2020open}. For datasets with pre-defined splits (ogb datasets, LRGB datasets, CLUSTER), we adopt the pre-defined splits. For datasets without pre-defined splits, we adopt a 8:1:1 train:validation:test random split for graph-level datasets (QM9 and QM8) and a 2:2:6 random split for node-level datasets (Citeseer and Pubmed). For QM9 and QM8, We follow the guidelines in MoleculeNet \cite{wu2018moleculenet} for choosing regression tasks and metrics. We perform joint training on 12 tasks for QM9 and 16 tasks for QM8.

We compare our method against (1) GNNs include GCN \cite{kipf2016semi}, GCNII \cite{chen2020simple}, GAT \cite{velivckovic2017graph}, GIN (GINE) \cite{xu2018powerful,hu2019strategies}, GatedGCN \cite{bresson2017residual} and APPNP \cite{klicpera2018predict}; (2) graph Transformers include GT \cite{dwivedi2020generalization}, SAN \cite{kreuzer2021rethinking}, Gophormer \cite{zhao2021gophormer}, SAT \cite{chen2022structure}, GraphGPS \cite{rampavsek2022recipe} and ANS-GT \cite{zhang2022hierarchical}. Most of the baseline results are cited from their original pepers, except for the results on QM9 and QM8 dataset and results of SAT on LRGB datasets. The more detailed settings for baselines are listed in the appendix. All our experiments are performed on one NVIDIA RTX 4090 GPU.

\subsection{AnchorGT Achieves Full Attention Performance}

\begin{figure*}[!ht]
\centering
\begin{minipage}{0.60\columnwidth}
\hspace{-2cm}
\begin{tikzpicture}
\begin{axis}[
    axis lines = left,
    ylabel=GPU Memory (GB),
    xlabel=Graph Size $n$,
    xmin=200, xmax=3300,
    ymin=0, ymax=22,
    width=7cm,height=6cm,
    legend style={at={(0.2,0.91)},nodes={scale=0.8, transform shape},anchor=west},
]
\addplot[color=red,mark=*,mark size = 0.7pt] coordinates {(500, 0.7) (1000, 1.8) (1500, 3.9) (2000, 8.3) (2500, 13.2) (3000, 18.9)};
\addplot[color=blue,mark=*,mark size = 0.7pt] coordinates {(500, 0.49) (1000, 1.10) (1500, 2.23) (2000, 3.20) (2500, 4.60) (3000, 6.39)};

\legend{GraphGPS,GraphGPS-AnchorGT}
\end{axis}
\end{tikzpicture}
\subcaption{GPU Memory Consumption}
\label{fig:sub1}
\end{minipage}\hspace{1cm}
\begin{minipage}{0.60\columnwidth}
\hspace{-1cm}
\begin{tikzpicture}
\begin{axis}[
    axis lines = left,
    ylabel=Epoch Time (s),
    xlabel=Graph Size $n$,
    xmin=200, xmax=3300,
    ymin=0, ymax=2.8,
    width=7cm,height=6cm,
    legend style={at={(0.2,0.91)},nodes={scale=0.8, transform shape},anchor=west},
]
\addplot[color=red,mark=*,mark size = 1pt] coordinates {(500, 0.23) (1000, 0.46) (1500, 0.71) (2000, 1.07) (2500, 1.63) (3000, 2.55)};

\addplot[color=blue,mark=*,mark size = 1pt] coordinates {(500, 0.20) (1000, 0.41) (1500, 0.67) (2000, 0.95) (2500, 1.32) (3000, 1.70)};

\legend{GraphGPS,GraphGPS-AnchorGT}
\end{axis}
\end{tikzpicture}
\subcaption{Time of Each Training Epoch}
\label{fig:sub2}
\end{minipage}

\hspace{1cm}
    
\caption{GPU memory consumption and training epoch time of synthetic efficiency tests on GraphGPS and its AnchorGT variant.}
\label{fig3}
\end{figure*}
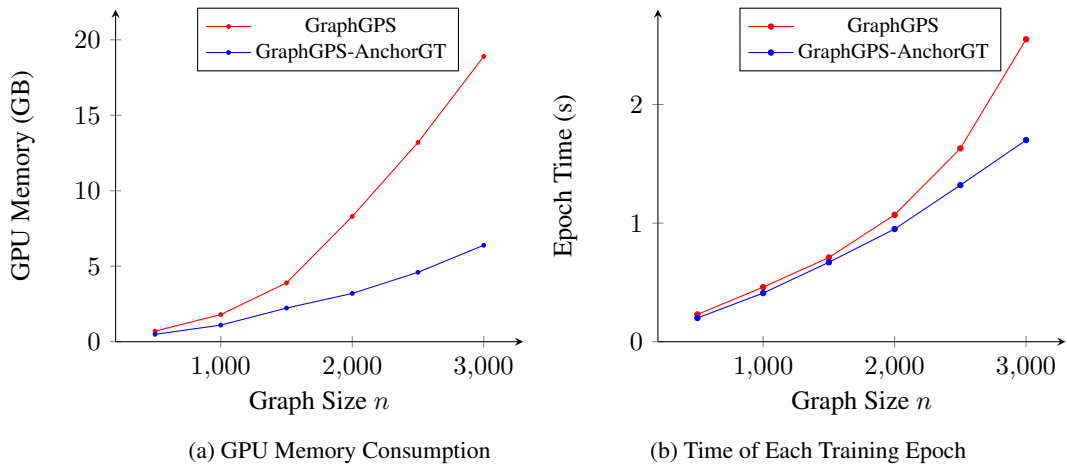

First, we analyze how the AnchorGT attention affects the performances of graph Transformer models on both graph-level and node-level benchmarks.  We select two important graph Transformer models for graph representation learning: Graphormer \cite{ying2021transformers} and GraphGPS \cite{rampavsek2022recipe}, and build their AnchorGT variants. Graphormer generally improves the basic graph Transformer with three kinds of structural encodings: centrality encoding, spatial encoding and edge encoding. Fortunately, the above three types of encodings can be realized within the framework of our anchor-based attention mechanism, so we directly constructed Graphormer-AnchorGT as the AnchorGT variant of Graphormer. Meanwhile, the GraphGPS architecture includes structural encoding, a local message passing module implemented by GNN, and a global attention mechanism module implemented by Transformer network. The global attention module adopts the standard Transformer structure or other linear Transformers, thus we can easily replace it with the anchor-based attention mechanism to construct GraphGPS-AnchorGT, the basic AnchorGT variant of GraphGPS. Additionally, considering the lack of relative structural encoding in Transformer networks of GraphGPS, we construct GraphGPS-AnchorGT-SPD as an enhanced version of GraphGPS-AnchorGT with SPD relative structural encoding to study the effect of structural encodings. For node-level tasks, we selected ANS-GT \cite{zhang2022hierarchical}, which achieves state-of-the-art and enhances the performance of graph Transformers in node-level tasks through novel sampling and attention methods. The Transformer component of ANS-GT is similar to that of Graphormer, with only a few specific requirements (we need to set the center node of each sampling as the anchor node). Therefore, we built the ANS-GT-AnchorGT model based on the previous approach. For all models, we adopt configurations specified in the original papers, and conduct a hyperparameter search on some training parameters (e.g., learning rate, batch size) to obtain the optimal configuration of AnchorGT models. We set $k=2$ for all AnchorGT models..

We summarize the results in Table \ref{tbl1}, \ref{tbl2} and \ref{tbl3}. For Graphormer and GraphGPS, the inclusion of anchor-based attention results in generally improved performance on some datasets. And for GraphGPS, SPD structure encoding improves the model's structural expressivity and boosts performance, resulting in slightly better performance on five datasets for the GraphGPS-AnchorGT-SPD model compared to the GraphGPS. This corresponds to our theoretical analysis in Section \ref{sec5}, that the expressive power of graph Transformer blocks can be improved by adding structural encodings, and AnchorGT layers with SPD are theoretically more powerful than GNN and naive Transformer layers. And for node-level tasks with ANS-GT, AncorGT also results in performance improvements, which can be attributed to the fact that anchor-based attention can reduce long-range noises in large graphs.

\subsection{AnchorGT is Fast and Memory Efficient}

Next, we investigate AnchorGT's ability to improve model efficiency, in terms of both GPU memory utilization during training and time of each training epoch. To run a standarized benchmark and avoid structural noises in real-world datasets, we construct a synthetic random graph dataset to characterize the efficiency of the model. We generate a training dataset consisting of 100 Erdos-Renyi random graphs, where the probability of an edge is 0.0001 (approximately reflecting the sparsity density of real-world graph datasets), and the number of nodes in each graph is controlled by a parameter $n$. A GraphGPS model consisting of only an input feature encoder and Transformer units with a hidden layer dimension of 128 and 5 Transformer blocks with 4 attention heads is constructed as the baseline. The batch size is set to 10. The anchorGT version of the model is kept consistent and k is set to 2. After warm-up, we measure the peak memory utilization using the built-in \texttt{get\_summary} function of \texttt{pytorch.cuda}and calculate the time of each training epoch.

We plot the results in Figure \ref{fig3}. As shown in the figure, the standard Transformer model's memory consumption exhibits a quadratic increase with the increasing number of graph size, limiting the scalability of the model. However, our AnchorGT model's memory consumption exhibits near-linear growth, maintaining low memory consumption as the graph size increases, reducing memory consumption by about $60\%$ compared to the baseline model. Therefore, our approach of saving memory consumption without sacrificing performance is meaningful for graph Transformer models. For training time, due to the extra computation introduced by the anchor method, AnchorGT saves around $10-30\%$ of the time compared to the original model. In summary, this synthetic test well demonstrates that AnchorGT is fast and memory efficient.

\subsection{Effect of $k$}
\label{sec64}
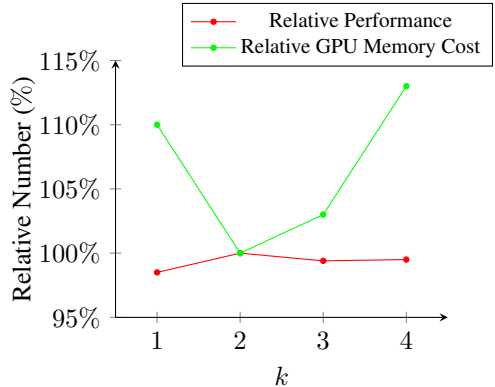
\begin{figure}[!ht]
\centering
\begin{minipage}{0.60\columnwidth}
\centering
\begin{tikzpicture}
\begin{axis}[
    axis lines = left,
    ylabel=Relative Number ($\%$),
    xlabel=$k$,
    xmin=0.5, xmax=4.5,
    ymin=95, ymax=115,
    width=6cm,height=5cm,
    yticklabel={$\pgfmathprintnumber{\tick}\%$},
    legend style={at={(0.2,1.1)},nodes={scale=0.8, transform shape},anchor=west},
]
\addplot[color=red,mark=*,mark size = 1pt] coordinates {(1, 98.5) (2, 100) (3, 99.4) (4, 99.5)};

\addplot[color=green,mark=*,mark size = 1pt] coordinates {(1, 110) (2, 100) (3, 103) (4, 113)};

\legend{Relative Performance,Relative GPU Memory Cost}
\end{axis}
\end{tikzpicture}
\end{minipage}
\hspace{1cm}
    
\caption{Relative Performance and Memory Cost ($\%$) of GraphGPS-AnchorGT-SPD on qm9 dataset. The settings stay the same with Table \ref{tbl2}.}
\label{fig4}
\end{figure}

Finally, we study a key parameter $k$ in the model, which determines the size of the neighborhood covered by the anchor points. Generally speaking, the larger $k$ is, the larger the coverage range of each anchor point, and the fewer the total number of anchor points. Specifically, the theoretical complexity of the model is $O(N(n_k+A))$, where $n_k$ increases with the increase of $k$, but $A$ decreases. Thus, the influence of parameter $k$ on the computational complexity and performance of the model is complex. In Table \ref{tbl4}, we present the number of anchors for the QM9 and ogbn-products datasets, and in Figure \ref{fig4} we plot the relative memory usage and performance of the GraphGPS-AnchorGT-SPD model on the QM9 dataset under different $k$ values. As depicted in Figure \ref{fig4} \footnote{Note that ogbn-products is a singlue large graph and qm9 consists of 130k small graphs, so the overall $N, E, |A|$ are reported for ogbn-products and the average $N,E,|A|$ of all graphs are reported for qm9. OOM stands for out of memory when calculating.}, the value of k has an impact on both the performance and complexity of the model, and the model exhibits the best performance and efficiency when $k=2$. In general, we found the following conclusions in experiments: $k=1$ is not a good choice, which will lead to too many anchor points, increase the complexity of the model, and reduce performance; when $k\geq 3$, the $k-$hop neighborhood of each node is also too large and this will hurt performance and efficiency. So in general, taking $k=2$ can achieve good performance and complexity at most circumstances.

\section{Conclusion}
In our study, we have presented AnchorGT, a highly efficient and adaptable attention architecture for graph Transformers. AnchorGT's versatility enables it to seamlessly integrate with a variety of graph Transformers, thereby significantly enhancing the scalability of the model, all while maintaining optimal performance levels.

\clearpage

\section*{Acknowledgements}
This work was supported by the National Natural Science Foundation of China (Grant No. 62276006).

\bibliographystyle{named}
\bibliography{ijcai23}

\clearpage
\appendix

\section{Proofs}
\subsection{Remark on Randomness in AnchorGT}
Prior to commencing the theoretical analysis below, it is necessary to highlight the issues of randomness present in the model. Previous simple GNN models and graph Transformer models are \textit{deterministic} algorithms, meaning they guarantee the same output for multiple runs on the same input graph. However, our method incorporates \textit{randomness} in the selection of anchors, meaning that for the same input graph, the anchor selection algorithm may yield different selections, resulting in the possibility of the model producing different outputs for the same input graph. Therefore, we apply the widely accepted definition established in previous research on random GNNs (e.g. DropGNN \cite{papp2021dropgnn}) below:
\begin{definition}[Discriminative Power of Randomized Graph Models]
    A model $M$ can distinguish between graphs $G_1$ and $G_2$ when it generates representations with different distributions for $G_1$ and $G_2$. 
\end{definition}
Here, the output representation is considered as a random variable. It can be observed that, since deterministic models produce a fixed representation for an input graph, this definition applies to both deterministic and randomized graph models. This definition will serve as the foundation for our subsequent theoretical discussion.

\subsection{Proof for Fact 1}
\begin{proof}

As the structural encoding function $\text{SE}(\cdot,\cdot)$ is neighbor-distinguishable, we can make embedding function $\mF$ satisfy:
\begin{align}
    &\mF(\text{SE}(v,u))=0, \text{ if }(v,u)\in E,\\
    &\mF(\text{SE}(v,u))=-\infty, \text{ if }(v,u)\notin E,
\end{align}
In Appendix A.2 and A.3 of Graphormer paper \cite{ying2021transformers} it is proved that by using such structural encodings, the Transformer layer can represent GNN layers and the readout function. Therefore, Fact 1 is proved.
\end{proof}

\subsection{Proof for fact 2}
\begin{proof}

Since the structural encoding function $\text{SE}(\cdot,\cdot)$ is neighbor-distinguishable, according to Fact 1 we can make the first Transformer layer represent a local GNN with SUM aggregation. Suppose after this step, the model yields $\vh_v=3$ for all 3-degree nodes, and $\vh_v=2$ for all 2-degree nodes. In the next layer, as the structural encoding function $\text{SE}(\cdot,\cdot)$ is both neighbor-distinguishable and anchor-distinguishable, we can make embedding function $\mF$ satisfy:
\begin{align}
    &\mF(\text{SE}(v,u))=a, \text{ if }(v,u)\in E\text{ and }u\notin A\\
    &\mF(\text{SE}(v,u))=b, \text{ if }u\in A,\\
    &\mF(\text{SE}(v,u))=-\infty, \text{ else},
\end{align}
where $a,b$ are two arbitrary numbers. Then after setting $\mP_Q,\mP_K$ to be zero mapping and $\mP_V$ to be identity mapping, the anchor-based attention will produce the following result:
\begin{align}
    \vh_v&=\frac{|N_k(v)|e^a}{|N_k(v)|e^a+|\tilde S|e^b}\text{MEAN}(\{\vh_u:u\in N_k(v)\})\\
    &+\frac{|\tilde S|e^b}{|N_k(v)|e^a+|\tilde S|e^b}\text{MEAN}(\{\vh_a:a\in S\}),
\end{align}

\begin{figure}[t]
\centering
\begin{subfigure}[b]{0.25\textwidth}
  \includegraphics{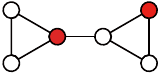}
  \subcaption{$G_1$}
\end{subfigure}%
\begin{subfigure}{.45\linewidth}
  \centering
  \includegraphics{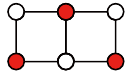}
  \subcaption{$G_2$}
\end{subfigure}
\caption{Two graphs in the proof for Fact \ref{fact2}.}
\label{fig2}
\end{figure}

where $\tilde S=S\setminus N_k(v)$. Given that the values of $a,b$ can be arbitrary, the value of $p=\frac{|N_k(v)|e^a}{|N_k(v)|e^a+|\tilde S|e^b}$ can be any number in the interval (0, 1). Suppose $\frac{e^a}{e^b}=\frac 12$, then it is trivial to verify that regardless of randomness, this attention layer will generate the node embedding set for $G_1$ and $G_2$ as follows:
\begin{align}
    &\ldblbrace\vh_v:v\in G_1\rdblbrace=\ldblbrace\frac{17}7,\frac{12}5,\frac{12}5,\frac{12}5,\frac{13}5,\frac 52\rdblbrace,\\
    &\ldblbrace\vh_v:v\in G_2\rdblbrace=\ldblbrace\frac73,\frac73,\frac73,\frac73,\frac{19}8,\frac{19}8\rdblbrace.
\end{align}
Therefore, after a MEAN or SUM readout function, the model will generate different embeddings for $G_1$ and $G_2$, which completes the proof.
\end{proof}

\section{Algorithm for Anchor Selection}
\begin{algorithm}[H]
    \caption{$k-$DS Anchor Selection Algorithm}
    \label{alg1}
    \begin{flushleft}
    \textbf{Input}: Input graph $G=(V,E)$.\\
    \textbf{Parameter}: $k$, which determines the number of hops that the anchor nodes cover in their neighborhood. \\
    \textbf{Output}: Anchor set $S$. \\
    \end{flushleft}
    \begin{algorithmic}[1] 
        \FOR {$v\in V$}
        \STATE Calculate $\text{deg}(v)$, the degree of $v$ in $G$.
        \ENDFOR
        \STATE $S\leftarrow \emptyset$.
        \STATE $V_{\text{left}}\leftarrow V$.
        \WHILE{$V_{\text{left}}$ is not empty}
        \STATE $a\leftarrow \argmax_{s}(\{\text{deg}(s):s\in V_{\text{left}}\})$.
        \STATE $a\leftarrow\text{RandomSelect}(a)$.
        \STATE $S\leftarrow S\cup \{a\}$.
        \STATE $V_{\text{left}}\leftarrow V_{\text{left}}\setminus\{a\}$.
        \FOR {$v\in N_k(a)$}
        \IF {$v\in V_{\text{left}}$}
        \STATE $V_{\text{left}}\leftarrow V_{\text{left}}\setminus\{v\}$.
        \ENDIF
        \ENDFOR
        \ENDWHILE
        \STATE \textbf{return} $S$.
    \end{algorithmic}
\end{algorithm}

\section{Experiment Settings}
In all experiments, the optimal parameter settings reported in the model paper are used whenever possible. For SAT, we adopt its reported setting on ZINC dataset to perform experiments on QM9, QM8, LRGB-Peptides-struct, LRGB-Peptides-func. For Graphormer \cite{ying2021transformers}, we use its reported best setting on ZINC dataset to perform experiments on QM9, QM8, LRGB-Peptides-struct, LRGB-Peptides-func, and Graphormer-AnchorGT follow this setting. For GraphGPS \cite{rampavsek2022recipe}, we use its reported best setting on ZINC dataset to perform experiments on QM9 dataset.

\end{document}